\documentclass{bmvc2k}

\title{How Effective is Pre-training of Large Masked Autoencoders for Downstream Earth Observation Tasks?}

\addauthor{Jose Sosa}{jose.sosa@uni.lu}{1}
\addauthor{Mohamed Aloulou}{mohamed.aloulou@ext.uni.lu}{1}
\addauthor{Danila Rukhovich}{danila.rukhovich@uni.lu}{1}
\addauthor{Rim Sleimi}{rsleimi@hydrosat.com}{2}
\addauthor{Boonyarit Changaival}{bchangaival@hydrosat.com}{2}
\addauthor{Anis Kacem}{anis.kacem@uni.lu}{1}
\addauthor{Djamila Aouada}{djamila.aouada@uni.lu}{1}

\addinstitution{
 The Interdisciplinary Centre for Security, Reliability and Trust (SnT)\\
 University of Luxembourg\\
 Luxembourg
}
\addinstitution{
 Hydrosat\\
 Luxembourg
}

\usepackage{booktabs}

\runninghead{Sosa, et al.}{Pretraining of Large MAE for EO Tasks}

\def\etal{\emph{et al}\bmvaOneDot}

\usepackage{float}

\begin{document}
\newgeometry{inner=16mm,outer=6mm,top=3mm,includehead,bottom=5mm,heightrounded}
\maketitle

\begin{abstract}
Self-supervised pre-training has proven highly effective for many computer vision tasks, particularly when labelled data are scarce. In the context of Earth Observation (EO), foundation models and various other Vision Transformer (ViT)-based approaches have been successfully applied for transfer learning to downstream tasks. However, it remains unclear under which conditions pre-trained models offer significant advantages over training from scratch. In this study, we investigate the effectiveness of pre-training ViT-based Masked Autoencoders (MAE) for downstream EO tasks, focusing on reconstruction, segmentation, and classification. We consider two large ViT-based MAE pre-trained models: a foundation model (Prithvi) and SatMAE. We evaluate Prithvi on reconstruction and segmentation-based downstream tasks, and for SatMAE we assess its performance on a classification downstream task. Our findings suggest that pre-training is particularly beneficial when the fine-tuning task closely resembles the pre-training task, e.g. reconstruction. In contrast, for tasks such as segmentation or classification, training from scratch with specific hyperparameter adjustments proved to be equally or more effective.
\end{abstract}
\restoregeometry

\section{Introduction}
\label{sec:intro}

Self-supervised learning has been widely used for NLP \cite{devlin2018bert,brown2020language,vaswani2017attention} and subsequently for computer vision tasks \cite{mahajan2018exploring,he2022masked,feichtenhofer2022masked,musallam2024self,karadeniz2024picasso,sosa2023self}. 
This practice aims to learn robust representations through pre-training models on large amounts of unlabelled data, followed by fine-tuning on particular downstream tasks where labels are available \cite{hendrycks2019using}. Initial successful attempts to apply the pre-training fine-tuning paradigm in the visual domain focused on various pre-training tasks \cite{he2022masked, he2020momentum, assran2023self}. However, thanks to the introduction of Vision Transformers (ViT) \cite{dosovitskiy2020image} and its posterior use for Masked Autoencoders (MAE) \cite{he2022masked}, the reconstruction task becomes a typical option for pre-training \cite{singh2023effectiveness,feichtenhofer2022masked,tong2022videomae,girdhar2023omnimae}.

Pre-training and subsequent fine-tuning of large ViT-based MAEs require significant computing resources \cite{mendieta2023towards}. Although this cost is justified for standard tasks like classification on datasets such as ImageNet \cite{deng2009imagenet}, the benefits in other contexts, like the medical domain \cite{khan2023revisiting} and Earth Observation (EO) \cite{corley2024revisiting, dionelis2024evaluating}, are less clear. In the case of EO, studies typically compare the performance of fine-tuning pre-trained ViT-based models against training from scratch well-known backbones such as ResNet \cite{he2016deep}, ConvViT \cite{gao2022convmae}, and U-Net \cite{chen2017rethinking}. However, these studies often lack rigorous justification for the performance gains resulting from pre-training, particularly missing basic hyperparameter tuning experiments \cite{jakubik2023foundationmodelsgeneralistgeospatial, satmae2022, noman2024rethinking}. This raises questions about the cost-effectiveness of pre-training for downstream tasks.

In this study, we investigate the effectiveness of relying on pre-trained large ViT-based MAEs for downstream tasks in the EO domain. Our objective is to determine whether training models from scratch for specific downstream tasks, with certain hyperparameter adjustments, can match or surpass the performance of initialising these from pre-trained large ViT-based MAEs. For this analysis, we focus on Prithvi \cite{jakubik2023foundationmodelsgeneralistgeospatial}, a foundation model that has been successfully applied to various segmentation and reconstruction tasks. Additionally, we analyse SatMAE \cite{satmae2022}, another large MAE ViT-based model designed for classification tasks. Although SatMAE is not considered as a foundation model, its structural similarity to Prithvi (both built on the ViT-based MAE architecture) makes it suitable for comparison in our study. We maintain the original distinction between the models, using Prithvi \cite{jakubik2023foundationmodelsgeneralistgeospatial} for segmentation and reconstruction tasks, and SatMAE \cite{satmae2022} for classification.

\section{Related Work}
\label{sec:rw}

Self-supervised pre-training of ViT-based models, particularly MAE, has proven beneficial in general settings \cite{he2022masked,feichtenhofer2022masked,xu2021vitae,zhang2023vitaev2} relying on standard datasets such as ImageNet \cite{deng2009imagenet}. Consequently, this approach has been widely explored in other specific domains, including the medical field \cite{xiao2023delving,khan2023revisiting} and, most relevant to our study, EO \cite{jakubik2023foundationmodelsgeneralistgeospatial, satmae2022,noman2024rethinking, reed2023scale, wang2022empirical}. The vast amount of unlabelled data available for EO and remote sensing has enormously benefited the scaling of pre-training MAEs and other ViT-based models \cite{kaplan2020scaling}, encouraging the proliferation of many foundation models \cite{bommasani2021opportunities}, such as Prithvi \cite{jakubik2023foundationmodelsgeneralistgeospatial}, SpectralGPT \cite{hong2024spectralgpt}, S2MAE \cite{li2024s2mae}, and SkySense \cite{guo2024skysense}. However, pre-training foundation models for EO requires large volumes of data and computing power \cite{mendieta2023towards}, restricting their development to well-resourced research groups \cite{lacoste2024geo, kolides2023artificial}. Therefore, pre-traning of similar models but with less parameters and `smaller' datasets have also been popular choice for transfer learning. Examples include SatMAE \cite{satmae2022}, ScaleMAE \cite{reed2023scale}, Cross-Scale MAE \cite{tang2024cross}, and SatMAE++ \cite{noman2024rethinking}.

In parallel to the rise of foundation models for computer vision tasks \cite{zou2024segment,kirillov2023segment,oquab2023dinov2,radford2021learning,li2022blip,yuan2021florence}, many studies have also surged analysing their capabilities for transfer learning on downstream general domain tasks \cite{yuan2023power, kolides2023artificial}, as well as for specialised domains \cite{huix2024natural, zhang2023challenges}. For example, Huix \etal \cite{huix2024natural} evaluate the performance of popular foundation models such as SAM \cite{kirillov2023segment}, SEEM \cite{zou2024segment}, and DINO \cite{oquab2023dinov2} on multiple medical datasets, finding that not all foundation models are suitable for transfer learning to downstream tasks in the medical domain. Similarly, Zhang \etal \cite{zhang2023challenges} conduct an in-depth analysis of the opportunities and challenges of using large pre-trained models in the medical field. More broadly, Chen \etal \cite{chen2021empirical} and later Touvron \etal \cite{touvron2022three} introduce valuable studies related to pre-training and fine-tuning of ViTs. Their analyses offer insights into different combinations of hyperparameters that make the training of ViT-based models more stable and efficient.

Unlike the medical and general domains, unfortunately, there is still a lack of comprehensive studies analysing the pre-training of large models for EO \cite{wang2022empirical,ambrozio2023agenda,mai2023opportunities}. One of the few such analyses is by Wang \etal \cite{wang2022empirical}, which evaluates the benefits of pre-training models with ImageNet for EO downstream tasks. Although this approach shares some similarities with ours, it differs in the nature of pre-training. Specifically, our study focuses on models that utilise domain-specific datasets during the pre-training stage, rather than relying on general-purpose datasets like ImageNet.


\section{Proposed Study}

\label{sec:data-cons}
\begin{figure}[ht]
    \centering
    \includegraphics[width=\textwidth]{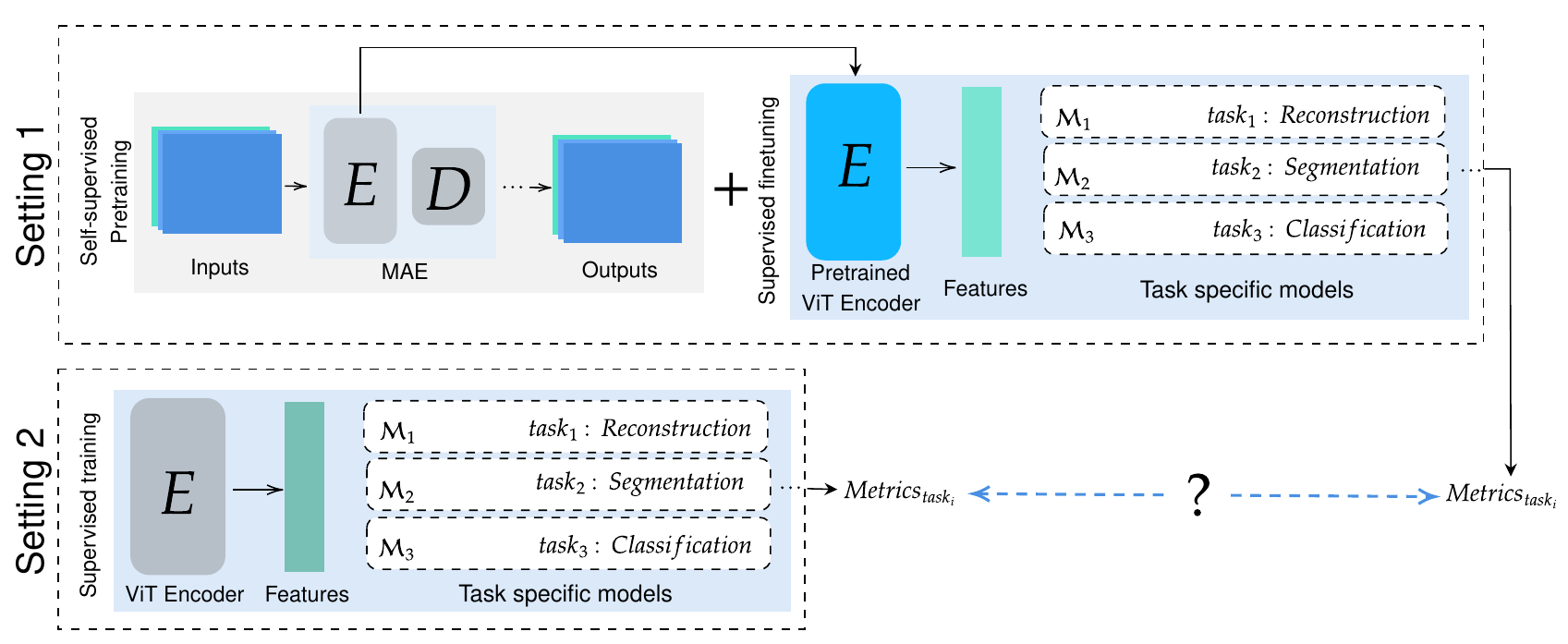}
    \caption{Settings for evaluation of ViT-based models. Setting 1 indicates that the encoder $E$ coupled with $M_i$ has been initialised with pre-trained weights. Setting 2 denotes the same, but without relying on pre-trained weights for $E$. Task related metrics have been compared for both settings to assess the effect of the self-supervised pre-training stage.}
    \label{fig:mae-vit}
\end{figure}

Our study investigates the effectiveness of pre-training large ViT-based MAE models for downstream EO tasks. We analyse two settings, as illustrated in \autoref{fig:mae-vit}. \textbf{Setting 1} involves initialising the encoder $E$ with pre-trained weights obtained from a self-supervised pre-training stage. The encoder $E$ is then coupled with a task-specific model $M_i$ and fine-tuned using supervised learning. In \textbf{Setting 2}, the self-supervised pre-training stage is omitted, and $E$ plus $M_i$ are trained from scratch. We compare the corresponding task-specific metrics for both settings.

Due to the structural similarities between the models and the diverse datasets used for their pre-training, we choose the encoders $E$ from Prithvi \cite{jakubik2023foundationmodelsgeneralistgeospatial} and SatMAE \cite{satmae2022} for our experimental settings. Specifically, for Prithvi, we analyse its performance in the reconstruction and segmentation tasks outlined in \cite{jakubik2023foundationmodelsgeneralistgeospatial}, including temporal cloud gap imputation, temporal crop segmentation, flood mapping, and wildfire scar mapping. Additionally, we evaluate the robustness of the features learnt from cloud imputation fine-tuning when applied to crop segmentation. For SatMAE, we follow the original implementation and evaluate it exclusively on the classification task.

Note that for obtaining the results reported throughout \autoref{sec-4}, all our experiments follow \textbf{Setting 2}. In other words, to build the models, we take the encoder $E$ either from Prithvi or SatMAE, without initialisation, and couple it with a task-specific model $M$. Then, we perform supervised training of $E$ and $M$ several times with different sets of hyperparameters and report the corresponding metrics. In the case of \textbf{Setting 1}, since it represents standard pre-training and finetuning, we simply rely on metrics from the related original implementations. Experiments corresponding to \textbf{Setting 1} will be normally indicated with the name of the model used for initialising $E$ (Prithvi or SatMAE), while results for \textbf{Setting 2} will be denote either as `scratch' or `scratch + hyp'.

\vspace{0.2cm}
\noindent \textbf{Data and Considerations.} We categorise our experiments into three main tasks: reconstruction, segmentation, and classification. Accordingly, we utilise different collections of data for each task. For reconstruction, we use data from the Multi-Temporal Cloud Gap Imputation dataset \cite{hls-multi-temporal-cloud-gap-imputation}. For segmentation, we utilise the Multi-Temporal Crop Segmentation dataset \cite{hls-multi-temporal-crop-classification-model}, Sen1Floods11 \cite{Bonafilia_2020_CVPR_Workshops}, and Wildfire Scar Mapping \cite{Prithvi-100M-burn-scar}. In the classification task, we rely on data from EuroSAT \cite{helber2019eurosat}. Given the high computational cost of experimenting with all possible combinations of hyperparameters for \textbf{Setting 2}, particularly when choosing ViTs for $E$, we strategically select key hyperparameters following some ideas from \cite{chen2021empirical}. We focus on general hyperparameters like learning rate, learning rate scheduler, and batch size, as well as ViT-specific ones such as the number of heads and layers. To determine the optimal configuration, we experiment using small data subsets with various hyperparameter combinations selected for fixed ranges of values. Note that we consistently use a multistep learning rate (MultiStepLR) scheduler across all experiments, with decay occurring after 67\% and 92\% of the total number of epochs. A ViT-Large is also fixed as the backbone for most experiments, unless otherwise specified. For all other hyperparameter values we provide details in the proper following sections.

\section{Results and Discussion} 
\label{sec-4}

\subsection{Multi-Temporal Cloud Gap Imputation}
\label{sec-clou-filling}

\begin{figure}[ht]
    \centering
    \includegraphics[width=\textwidth]{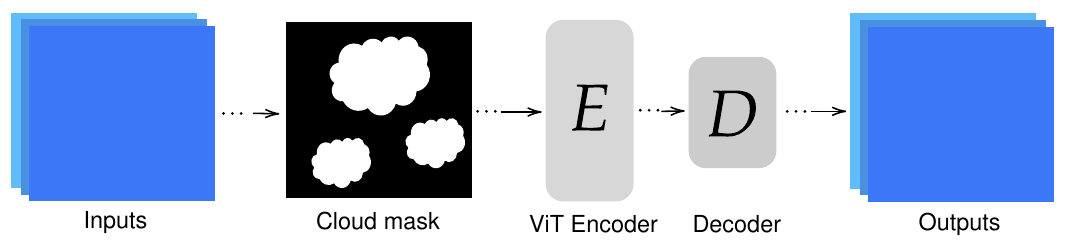}
    \caption{Training MAE for cloud imputation. Unlike the standard MAE which relies on fixed masking ratio, in this case we provide binary cloud masks for the inputs to train $E$ and $D$ from scratch.}
    \label{fig:finetuning-clou-imp}
\end{figure}

For the task cloud gap imputation, we rely on the approach proposed in \cite{jakubik2023foundationmodelsgeneralistgeospatial}. In general, the task is simply image reconstruction, involving the use of a MAE to reconstruct regions covered by clouds in the given input image as illustrated in \autoref{fig:finetuning-clou-imp}. Note that in this case, the task for fine-tuning is exactly the same as that for pre-training. However, unlike the standard MAE masking \cite{he2022masked}, we follow \cite{jakubik2023foundationmodelsgeneralistgeospatial} and use the binary cloud masks in the dataset \cite{hls-multi-temporal-cloud-gap-imputation} to build the masks needed for the inputs. 


Following \textbf{Setting 2}, we train from scratch the components $E$ and $D$ (where $D$ correspond to the task-specific model) of the model depicted in \autoref{fig:finetuning-clou-imp}. We use the same $E$ and $D$ architectures as in Prithvi and train these with data from the multi-temporal cloud imputation dataset \cite{hls-multi-temporal-cloud-gap-imputation}. We perform several experiments with different combinations of hyperparameters. In summary, we lower the starting learning rate to $0.00005$, replace the scheduler as mentioned in \autoref{sec:data-cons}, and reduce the number of heads and layers in the ViT encoder to 3 and 2, respectively. We train the model for 200 epochs, while maintaining the same batch size, patch size, and other hyperparameter values as indicated in \cite{jakubik2023foundationmodelsgeneralistgeospatial}. We further investigate the impact of changing the encoder backbone for the MAE. In addition to the original ViT-Large backbone, we experiment with ViT-Base and ViT-Small \cite{dosovitskiy2020image}. Following \cite{jakubik2023foundationmodelsgeneralistgeospatial}, we evaluate all configurations using the test set from \cite{hls-multi-temporal-cloud-gap-imputation}. \autoref{table:cloud-comp} reports the results in terms of mean absolute error (mae\footnote{To avoid confusion with Masked Autoencoder (MAE), mean absolute error is denoted in lowercase.}) and structural similarity index (SSIM).

\begin{table}[h]
\centering
\begin{tabular}{l ccc}
\toprule
\textbf{Initialisation} & \textbf{Backbone} & \textbf{mae}  & \textbf{SSIM} \\ 
\midrule
Prithvi       & ViT-Large  &\textbf{0.020}  & \textbf{0.972} \\ 
\midrule
Scratch + hyp  & ViT-Large  &0.025  & 0.964 \\ 
Scratch + hyp  & ViT-Base   & 0.025  & 0.964 \\ 
Scratch + hyp  & ViT-Small & 0.027  & 0.959 \\ \bottomrule
\end{tabular}
\caption{Comparison of evaluation for cloud gap imputation. The first column indicates the initialisation used for $E$. The next columns provide details on the backbones and evaluation metrics.}
\label{table:cloud-comp}
\end{table}

Although altering the encoder backbone $E$ results in a significant reduction in model parameters and accelerates the training time, it does not surpass the performance of the ViT-Large backbone used in \cite{jakubik2023foundationmodelsgeneralistgeospatial}. We hypothesise that since the cloud imputation task is identical to the pre-training task, training from scratch with hyperparameter tuning has a minimal impact on the final performance. In this context, initialisation with pre-trained weights from Prithvi provides a beneficial effect on fine-tuning, compared to training from scratch.

\subsection{Multi-Temporal Crop Segmentation}
\label{sec-crop-seg}

\begin{figure}[ht]
    \centering
    \includegraphics[width=\textwidth]{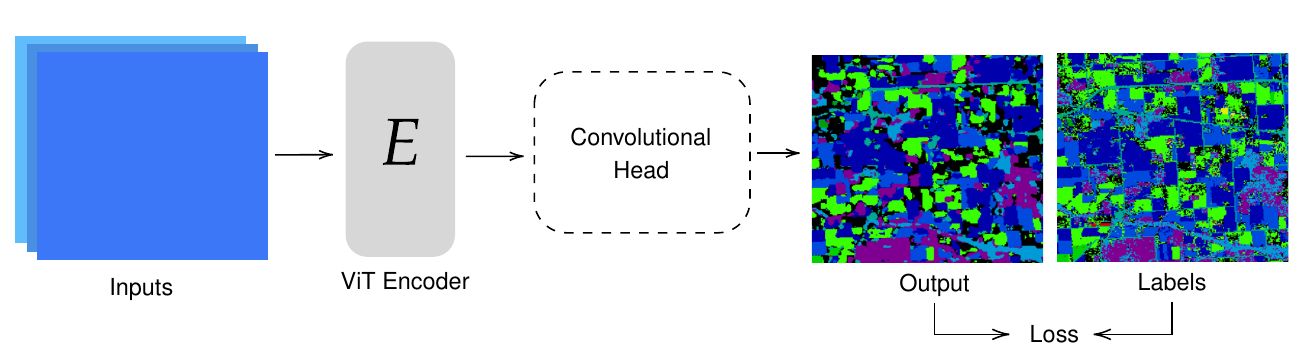}
    \caption{Standard emsemble for segmentation tasks. For crop classification, $E$ has been coupled with a convolutional head, which could contain convolutional and linear layers. Labels are provided on the dataset for supervised training of the model.}
    \label{fig:seg-model}
\end{figure}

For the task of crop segmentation, we rely on the architecture depicted in \autoref{fig:seg-model}, which consists of a ViT-based encoder $E$ coupled with a convolutional head. For training the model, we use labelled data from the multi-temporal crop segmentation dataset \cite{hls-multi-temporal-crop-classification-model} as in \cite{jakubik2023foundationmodelsgeneralistgeospatial}. 

\begin{figure}[ht!]
     \centering
     \includegraphics[width=\textwidth]{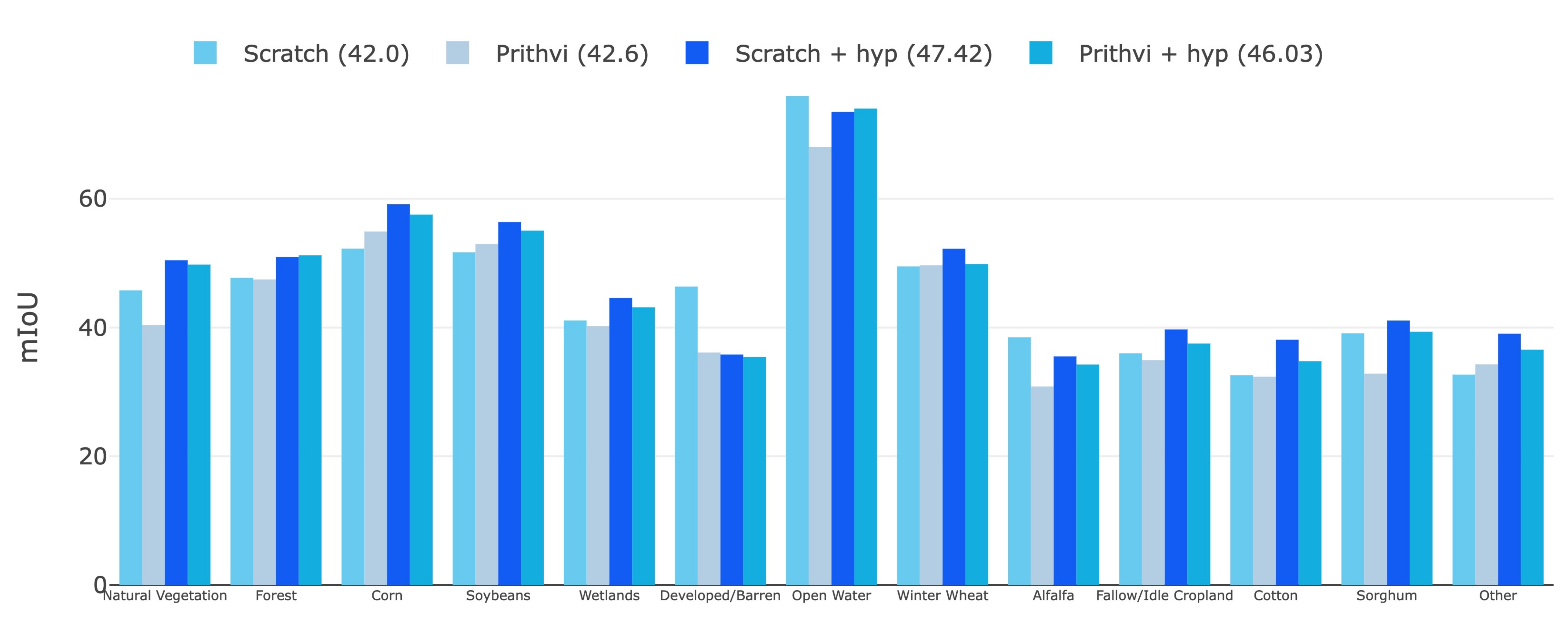}
     \caption{Comparison of performance of the model under different initialisations. We calculate the mIoU for each of the crop and land cover classes on the test set. Averages for each setting appear on the legend located at the top of the plot.}
     \label{fig:miou_seg_crop}
 \end{figure}

In line with \textbf{Setting 2}, we train the model from scratch with the same hyperparameters as training with Prithvi initialisation (Scratch) and with some hyperparameters adjustments (Scratch + hyp). In addition, we extend the analysis under \textbf{Setting 1} by exploring hyperparameters adjustments when initialising $E$ with Prithvi (Prithvi + hyp). \autoref{tab:params-cropseg} summarises the hyperparameters values used for each of these experiments. After training, we perform evaluation using the corresponding data from \cite{hls-multi-temporal-crop-classification-model}. \autoref{fig:miou_seg_crop} shows the mean Intersection over Union (mIoU) for each of the 13 crop and land cover classes in the test set. The average values for all settings are indicated in the legend at the top of the figure.

\begin{table}[hbt]
    \centering
    \begin{tabular}{lc c c cc}
    \toprule
         \textbf{Intialisation}  & \textbf{Frames}  & \textbf{Initial lr}   & \textbf{Layers} & \textbf{Heads} & \textbf{Scheduler}\\
    \midrule
        Prithvi  & 3  & $1.5e-5$  & 6 &8 & Polynomial\\
    
        Prithvi + hyp & 3  & $1e-4$ & 6&12 & MultiStepLR \\
        \midrule
        Scratch & 3  & $1.5e-5$ & 6 &8 & Polynomial\\
        Scratch + hyp & 3  & $1e-4$  & 6&12 & MultiStepLR \\
    \bottomrule
    \end{tabular}
    \caption{Summary of hyperparameters for different settings. The first column refers to the type of initialisation for the ViT encoder $E$, either from Prithvi \cite{jakubik2023foundationmodelsgeneralistgeospatial} or from scratch.} 
    \label{tab:params-cropseg}
\end{table}

According to \autoref{fig:miou_seg_crop}, the average mIoU for fine-tuning the model initialised with Prithvi's weights (Prithvi) is nearly the same as starting training from scratch (Scratch), with just a small difference of $0.6$. Surprisingly, adjusting some hyperparameters in the latter setting (Scratch + hyp) leads to a significant increase in the average mIoU from 42.0 to 47.42. Additionally, using a combination of specific hyperparameters and initialisation from Prithvi (Prithvi + hyp) yields an average mIoU of 46.03, which is higher than the baseline performance (Prithvi), but still lower than the initialisation from scratch with the adjusted hyperparameters (Scratch + hyp).

\vspace{0.2cm}
\noindent\textbf{Multi-temporal Crop Segmentation with Cloudy Data.} The above experiments demonstrate that fine-tuning Prithvi can improve the performance of crop segmentation. Experiments reported in (\autoref{sec-clou-filling}), showcase a better reconstruction of cloudy inputs with fine-tuning Prithvi on this task.  However, the benefits of combining these tasks remain underexplored. In particular, it is unclear how effectively models pre-trained for cloud imputation perform on crop segmentation tasks when dealing with cloudy inputs (which is common in EO data).

To investigate this, we replicate the crop segmentation experiments described above with cloudy inputs. In particular, we simulate cloudy conditions using fixed masking ratios and apply these to crop segmentation data. We use masking levels of 30\%, 60\%, and 90\% to retrain the crop segmentation model in \autoref{fig:seg-model}. For each masking ratio, we conduct experiments with different initialisation of the ViT encoder $E$: Prithvi + hyp, Scratch + hyp, and Prithvi + cloud finetuning. Note that the hyperparameters for the first two experiments are as specified in \autoref{tab:params-cropseg}, while the last simply follows the training described above for standard crop segmentation. 

\begin{figure}[H]
    \centering
    \includegraphics[width=\textwidth]{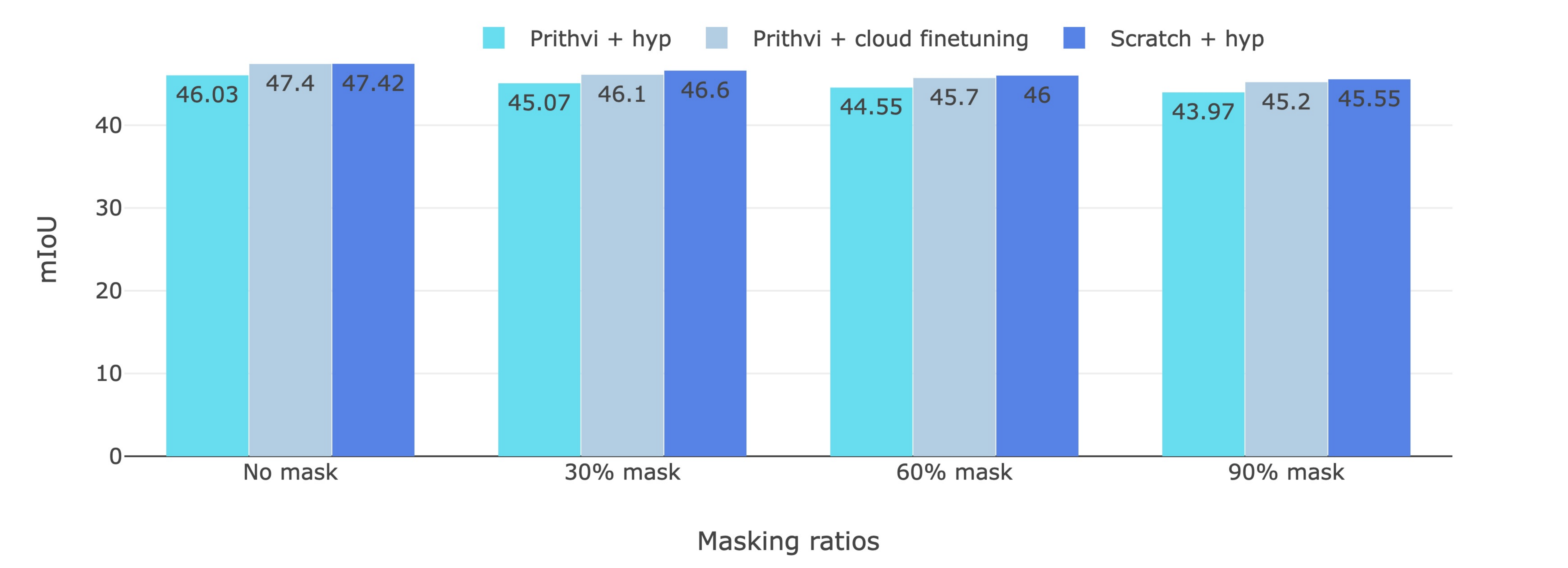}
    \caption{Different initialisation for crop segmentation with simulated cloudy data. Corresponding mIoU for each masking ratio with different initialisation for ViT Encoder.}
    \label{fig:masks-crops}
\end{figure}

As it can be observed from \autoref{fig:masks-crops}, initialising $E$ with pretrained Prithvi on cloud imputation (Prithvi + cloud finetuning) provides a slight improvement over standard Prithvi initialisation (Prithvi + hyp). However, contrary to what can be expected, training from scratch with hyperparameters adjustments (Scratch + hyp) is slightly better than other settings. Considering the extensive time and resources required for pre-training Prithvi, fine-tuning on the cloud imputation task, and subsequently fine-tuning again for crop segmentation, it is clear that using pretrained Prithvi is not a cost-effective initialisation in this scenario.


\subsection{Flood Mapping}
We extend the segmentation experiments to flood segmentation using the dataset from \cite{Bonafilia_2020_CVPR_Workshops}, and following \cite{jakubik2023foundationmodelsgeneralistgeospatial}. Unlike the data used for crop classification and cloud gap imputation, the Sen1Floods11 data set \cite{Bonafilia_2020_CVPR_Workshops} lacks a temporal dimension, which requires the segmentation model to work with individual images. However, the model maintains the same general structure as the one used for crop segmentation, featuring a ViT encoder coupled with a few convolutional layers (\autoref{fig:seg-model}).


The approach imitates the one described in \autoref{sec-crop-seg}, involving experiments under \textbf{Setting 2}, training the segmentation model from scratch plus some hyperparameters adjustments. \autoref{tab:flood_data} summarises the hyperparameters values used when training the model from scratch (Scratch + hyp). Note that in this case results when training from scratch (Scratch) with same hyperparameters as with Prithvi initialisation (Prithvi) are taken from the original paper.

\begin{table}[ht]
    \centering
    \begin{tabular}{lc c c ccc c}
    \toprule
        \textbf{Initialisation} &   \textbf{Frames} & \textbf{Batch} & \textbf{Initial lr}  & \textbf{Layers} & \textbf{Heads} &\textbf{Scheduler} \\
    \midrule
        Prithvi   & 1 & 4 & $1.5e-5$ & 12 & 12& Polynomial \\
        Scratch   & 1 & 4 & $1.5e-5$ & 12 &12 & Polynomial\\
        \midrule
        Scratch + hyp & 1 & 8 & $4e-5$  & 6 &6& MultiStepLR\\
    \bottomrule
    \end{tabular}
     \caption{Summary of hyperparameters for different experiments. First column denotes the type of initialisation for the ViT encoder $E$, either from Prithvi \cite{jakubik2023foundationmodelsgeneralistgeospatial} or from scratch.} 
    \label{tab:flood_data}
\end{table}

\begin{table}
    \centering
    \begin{tabular}{l lcc cc c}
        \toprule
        \textbf{Initialisation}& \textbf{Epochs} & \textbf{ IoU ($\uparrow$)} & \textbf{F1($\uparrow$)} & \textbf{mIoU($\uparrow$)} & \textbf{mF1($\uparrow$)} & \textbf{mAcc($\uparrow$)} \\
        \midrule
        Scratch & 50 & 80.67 & 89.30  & 88.76  & 93.85 & 94.79  \\
        Prithvi& 50  & 81.26 & 89.66 & 89.10 & 94.05 & \textbf{95.07}  \\
        Scratch& 500  & 82.97 & 90.69  & 90.14  & 94.66 & 94.82  \\
        Prithvi& 500 & 82.99 & 90.71 & 90.16 & 94.68 & 94.60   \\
        
        \midrule
        Scratch + hyp& 50 &81.2	&89.62&	89.1&	94.05&	94.84 \\
        
        Scratch + hyp& 100  & 82.15&	90.26&	89.73&	94.42&	94.93 \\

        Scratch + hyp& 400 & \textbf{83.11}	&\textbf{90.78}	&\textbf{90.24} &	\textbf{94.72} & 95.03  \\
        
        \bottomrule
    \end{tabular}
    \caption{Results for different initialisation of ViT encoder within segmentation model. Best results for each metric appear in bold.}
    \label{tab:flood-metrics}
\end{table}

We train the model from scratch for up to 50, 100, and 400 epochs and evaluate each of them using the test set from \cite{Bonafilia_2020_CVPR_Workshops}. We present results in \autoref{tab:flood-metrics}, including all metrics and results reported in \cite{jakubik2023foundationmodelsgeneralistgeospatial}. Similar to results with crop segmentation, few changes on the hyperparameters yield better performance than relying on Prithvi weights for initialisation. It is also worth noting that training from scratch significantly reduces the overall training time. Although this setting takes more epochs to match or surpass intialisation from pre-trained Prithvi in all the metrics, it is still more time efficient if we consider the fact that the Prithvi pre-training time is approximately 4.5 days \cite{jakubik2023foundationmodelsgeneralistgeospatial}.

\subsection{Wildfire Scar Mapping}
For wildfire scar segmentation experiments, we use data from the wildfire scar mapping dataset \cite{Prithvi-100M-burn-scar}. Following the same approach as for flood mapping, we train the model from scratch using the hyperparameters specified in the third row of \autoref{tab:params-burnscars}. 
\begin{table}[h]
    \centering
    \begin{tabular}{lc cc ccc}
    \toprule
         \textbf{Initialisation}  & \textbf{Frames} & \textbf{Batch} & \textbf{Initial lr}  & \textbf{Layers} & \textbf{Heads}& \textbf{Scheduler} \\
    \midrule
        Prithvi & 1 & 4 & $1.3e-5$  & 12 & 12 & Polynomial \\
        Scratch & 1 & 4 & $1.3e-5$  & 12 & 12 & Polynomial \\
        \midrule
        Scratch + hyp & 1 & 8 & $5e-5$  & 6&6 & MultiStepLR \\
    \bottomrule
    \end{tabular}
    \caption{Summary of hyperparameters for different experiments. First column refers to the type of initialisation for the ViT encoder $E$, either from Prithvi \cite{jakubik2023foundationmodelsgeneralistgeospatial} or from scratch.} 
    \label{tab:params-burnscars}
\end{table}

As shown in \autoref{tab:scar_burn_comparison}, training model from scratch for 100 epochs with some hyperpararemeters adjustments (Scratch + hyp) eventually outperforms all the metrics reported in \cite{jakubik2023foundationmodelsgeneralistgeospatial} for wildfire scar mapping (Prithvi and Scratch). Although the model trained from scratch requires twice as many epochs to surpass its counterpart's performance, it is still convenient when considering the time required for pre-training Prithvi.

\begin{table}[H]
    \centering
    \begin{tabular}{ll cccccc}
        \toprule
        \textbf{Initialisation} & \textbf{Epochs} & \textbf{IoU($\uparrow$)} & \textbf{F1($\uparrow$)} & \textbf{mIoU($\uparrow$)} & \textbf{mF1-score($\uparrow$)} & \textbf{mAcc($\uparrow$)}  \\
        \midrule
        Prithvi& 50 & 73.62 & 84.81 & 84.84 & 91.40 & 92.48 \\
        Scratch& 50 & 72.26  & 83.89  & 84.01  & 90.87  & 92.41  \\
        \midrule
        Scratch + hyp& 100 & \textbf{73.99} & \textbf{85.05} & \textbf{85.41} & \textbf{91.72} & \textbf{93.79}  \\
        \bottomrule
    \end{tabular}
    \caption{Performance comparison of different model initialisations for wildfire scar mapping.}
    \label{tab:scar_burn_comparison}
\end{table}

\subsection{Land Cover Classification}
In previous experiments we focus on reconstruction and segmentation tasks relying on Prithvi's encoder which is either initialised from scratch or pre-trained. Unlike typical pre-training approaches for large models, which usually rely on well-established datasets, the EO domain lacks a standardised dataset for pre-training. In the case of Prithvi, it has been pretrained with data from the NASA's HLS V2 L30 product \cite{claverie2018harmonized}. To demonstrate that our findings are not specific to any pretrained dataset or finetuning task, we utilise SatMAE, a structurally similar large ViT-based MAE model to Prithvi, but pre-trained with different data. In particular, we use the SatMAE encoder, pre-trained on data from \cite{christie2018functional}, for the land cover classification task. Following the same strategy as with segmentation experiments, we follow \textbf{Setting 2} to train from scratch a ViT model for classification with some hyperparameter adjustments (details of the model used could be found in \cite{satmae2022}). Specifically, we modify the initial learning rate to $6e-4$ and keep the MultiStepLR scheduler. We train and test the model using the EuroSAT dataset \cite{helber2019eurosat}. We experiment with both RGB and multispectral data, and report the top-1 accuracy in \autoref{tab:class_res_eurosat}, comparing the results with those of the model initialised with pre-trained ViT-encoder from SatMAE \cite{satmae2022} (\textbf{Setting 1}).


\begin{table}[hbt]
    \centering
    \begin{tabular}{llcc}
    \toprule
    \textbf{Initialisation}  & \textbf{Input} & \textbf{Epochs} & \textbf{Top-1 Acc ($\uparrow$)} \\
        \toprule
        SatMAE & RGB & 50  & 95.74  \\
        Scratch + hyp  & RGB& 50   & 95.78 \\
        Scratch + hyp  & RGB & 100& \textbf{97.00}  \\
        \midrule
        SatMAE  & Multi Spectral & 50 & \textbf{98.98}  \\
        Scratch + hyp   & Multi Spectral& 50 & 97.28  \\
        Scratch + hyp  & Multi Spectral & 100& 98.44  \\
        \bottomrule
    \end{tabular}
    \caption{Results for different initialisation of ViT encoder used for classification. Best results for each setting appear in bold.}
    \label{tab:class_res_eurosat}
\end{table}

Based on the results from \autoref{tab:class_res_eurosat}, training the model from scratch with RGB data yields better performance when compared to initialisation from pre-trained SatMAE. Notably, initialisation from Scratch + hyp outperforms the SatMAE initialisation, even when both are trained for the same number of epochs. Conversely, when using multispectral data, pre-training provides a slight improvement in performance.

\subsection{Discussion}
Based on the results from various EO segmentation and classification downstream tasks, we can observe that using large ViT-based MAE pre-trained models (\textbf{Setting 1}) does not consistently outperform models initialised from scratch (\textbf{Setting 2}). Our findings indicate that pre-training tends to improve performance for downstream tasks closely aligned with the pre-training task, such as the \textit{Multi-Temporal Cloud Gap Imputation} task. However, for most segmentation tasks—including \textit{Multi-Temporal Crop Segmentation, Multi-Temporal Crop Segmentation with cloudy data, Flood Mapping,} and \textit{Wildfire Scar Mapping}—initialisation from scratch, together with hyperparameter tuning, can achieve comparable or even superior results. Similarly, for land cover classification, \textbf{Setting 2} is beneficial when using RGB inputs. However, when using multi-sprectral data, initialisation from pretrained SatMAE shows better performance.

\section{Conclusion}
In this paper, we analyse the effectiveness of pre-training large ViT-based MAE models for downstream EO tasks, with focus on one foundation model (Prithvi) and SatMAE. We experiment on reconstruction, segmentation, and classification EO tasks, demonstrating that relying on large ViT-based MAE pre-trained models as initialisation does not consistently outperform models initialised from scratch. Given that our experiments involve a diverse range of datasets on finetuning and pre-training stages, we hypothesise that the limitations observed in pre-training MAE ViT-based models might be more related to model design than to the data itself. This suggests that better strategies for pre-training foundation models and other MAE ViT-based models for EO could enhance the benefits of the fine-tuning process for downstream tasks. However, it is important to note that this study is relatively small in scope. Future research should extend these findings by incorporating additional datasets and models, particularly for classification tasks.
\\
\\
\textbf{Acknowledgements.} This work is supported by FNR HPC BRIDGES project under the reference HPC\textunderscore BRIDGES/2022/17978225/AI4CC. The experiments were performed on the Luxembourg national supercomputer MeluXina. Thanks to LuxProvide teams for their support.


\bibliography{bmvc_review}
\end{document}